\documentclass{article}

     \usepackage[preprint]{neurips_2025}

\usepackage[utf8]{inputenc} 
\usepackage[T1]{fontenc}    
\usepackage{hyperref}       
\usepackage{url}            
\usepackage{booktabs}       
\usepackage{amsfonts}       
\usepackage{nicefrac}       
\usepackage{microtype}      
\usepackage{xcolor}         

\usepackage{array}
\usepackage{multirow}
\usepackage[]{subcaption}
\usepackage{algorithm}
\usepackage{algorithmic}
\usepackage{graphicx}
\usepackage{bbm}

\usepackage{xspace} 
\newcommand{\thesys}{\textsc{CLUE}\xspace}
\newcommand{\name}{\thesys}

\newcommand{\thesyss}{\textsc{CLUE}: Calibration via Learning Uncertainty–Error Alignment\xspace}
\newcommand{\nameAll}{\thesyss}

\title{CLUE: Neural Networks Calibration via Learning Uncertainty–Error alignment}

\author{%
   Pedro Mendes$^{1,2}$, Paolo Romano$^{2}$, David Garlan$^{1}$\\
    \textit{$^{1}$Software and Societal Systems Department, Carnegie Mellon University}\\
    \textit{$^{2}$INESC-ID and Instituto Superior Técnico, Universidade de Lisboa}   %
}

\begin{document}

\maketitle

\begin{abstract}
Reliable uncertainty estimation is critical for deploying neural networks (NNs) in real-world applications. While existing calibration techniques often rely on post-hoc adjustments or coarse-grained binning methods, they remain limited in scalability, differentiability, and generalization across domains. 
In this work, we introduce \name (Calibration via Learning Uncertainty–Error Alignment), a novel approach that explicitly aligns predicted uncertainty with observed error during training, grounded in the principle that well-calibrated models should produce uncertainty estimates that match their empirical loss. 
\name adopts a novel loss function that jointly optimizes predictive performance and calibration, using summary statistics of uncertainty and loss as proxies. The proposed method is fully differentiable, domain-agnostic, and compatible with standard training pipelines. 
Through extensive experiments on vision, regression, and language modeling tasks, including out-of-distribution and domain-shift scenarios, we demonstrate that \name achieves superior calibration quality and competitive predictive performance with respect to state of the art approaches without imposing significant computational overhead.
\end{abstract}

\section{Introduction}
\label{sec:intro}

Modern neural networks (NNs) often achieve high predictive accuracy but tend to produce poorly calibrated probability estimates~\cite{model_calibration}, i.e., their predicted confidence does not reliably reflect the true likelihood of correctness. This miscalibration can lead to overconfident or underconfident decisions, particularly problematic in critical applications such as medical diagnosis, autonomous driving, and risk-sensitive decision-making.
Accurate uncertainty estimation enables more reliable decision-making and trustworthy predictions. Ideally, a well-calibrated model should not only predict correctly but also express appropriate confidence in its predictions (e.g., reporting high confidence when it is likely correct and low confidence when it is incorrect).

Recent efforts that tackled this problem 
can be divided into two main categories: (i) post-processing methods, which adjust the model’s predicted probabilities~\cite{model_calibration,isotonic_regression,ece,uce,calib_regression1,calib_regression2,unc_lang} after training using a separate validation set; and (ii) training-time methods, which incorporate calibration objectives directly into the loss function to encourage accurate uncertainty estimation during learning~\cite{cals,euat,deep_ensembles,ce_pe,focal_loss,heteroscedastic1,heteroscedastic2}.
Post-hoc methods such as temperature scaling~\cite{soft_calibration} and isotonic regression~\cite{isotonic_regression} offer strong empirical performance with low computational overhead. However, they rely on the assumption that model predictions require only minor adjustments, which limits their effectiveness under severe miscalibration or poor uncertainty estimates. Their dependence on held-out validation data makes them sensitive to distribution shifts, and the need to tune additional calibration parameters further adds to their fragility. Moreover, these techniques only adjust the final output layer, without influencing the internal uncertainty representations, limiting their applicability in tasks that require uncertainty-aware reasoning throughout the entire model architecture.
In contrast, training-based approaches, while typically more computationally demanding, have shown superior performance in calibration and uncertainty estimation by incorporating uncertainty modeling directly into the learning process~\cite{focal_loss,focal_loss1,cals,euat}.

However, despite the increasing attention to uncertainty estimation in deep learning, existing methods~\cite{isotonic_regression,model_calibration,ce_pe,deep_ensembles,euat,cals,calib_regression1,calib_regression2} suffer from two key limitations. First, many approaches lack generalization across domains and architectures. For example, methods tailored to classification tasks are not readily applicable to regression, or generative modeling. Second, many uncertainty estimation techniques are computationally inefficient, 
making them costly to deploy in real-time or resource-constrained settings.

Thus, this work introduces \name (Calibration via Learning Uncertainty–Error alignment), a novel training method that improves uncertainty estimation in NNs by ensuring that predicted uncertainty aligns with the model’s error. Our approach addresses both domain generalization and computational efficiency. We promote alignment between the model uncertainty and error by matching summary statistics, such as predictive entropy (PE)~\cite{pe,pe_mi}, to the observed performance during training. Moreover, rather than relying on coarse binning-based approximations, we leverage the model’s loss as a proxy for the prediction error, providing a tractable, lightweight, and fully differentiable loss function.
Specifically, this work makes the following key contributions:

\textcolor{white}{n}\hspace{1mm}$\bullet$ \textit{\textbf{Generalizable Framework}}: We introduce a formulation for calibration that applies to NNs producing a predictive distribution, including classification, regression, and language models.

\textcolor{white}{n}\hspace{1mm}$\bullet$ \textit{\textbf{Uncertainty-Error alignment Calibration}}: We propose a scalable, differentiable loss function that aligns uncertainty with model's error by comparing summary statistics with loss-based proxies, eliminating the need for distributional comparisons or binning approximations.

\textcolor{white}{n}\hspace{1mm}$\bullet$ \textit{\textbf{Empirical Evaluation Across Domains}}: We demonstrate the effectiveness of our method on diverse benchmarks, including image and binary classification, probabilistic regression, language modeling, and out-of-distribution (OOD) detection.
   
\textcolor{white}{n}\hspace{1mm}$\bullet$ \textit{\textbf{Efficiency}}: We analyze the trade-off between calibration quality and computational cost, showing that our method achieves strong performance with a minimal number of MC dropout samples. 


\section{Background}

In the current section, we formally define the problem of correctly estimating the uncertainty of NNs and review and discuss existing methods to quantify the uncertainty of NNs.

\subsection{Problem Formulation}
\label{sec:problem}

Intuitively, the goal of calibrating neural models is to ensure that the predictive uncertainty of a model aligns with its true probability of correctness. 
There exist multiple mathematical formulations of calibration in the literature. One of the most widely adopted~\cite{model_calibration,ece} defines a classifier $f$ as perfectly calibrated if 
\begin{equation}
    \mathbb{P}(\hat{y}=y | \hat{p}=p)=p,  \hspace{3mm} p\in [0,1]
    \label{eq:calibration}
\end{equation}
where $\hat{y}$ is the model prediction, $y$ is the true class label, and $\hat{p}$ is the model's prediction confidence. Intuitively, this means that among all predictions made with confidence $p$, the model should be correct $p$ percent of the time.
To quantify miscalibration, a common approach is to compute the expected deviation between accuracy and confidence
\begin{equation}
    \mathbb{E}[ \mathbb{P}(\hat{y}=y | \hat{p}=p)-p], \hspace{3mm} p\in [0,1].
\end{equation}
This is often approximated via the Expected Calibration Error (ECE)~\cite{ece}, which bins predictions by confidence and aggregates the gap between empirical accuracy and average confidence per bin, i.e., 
\begin{equation}
    ECE=\sum_{m=1}^{M} \frac{|B_m|}{n} |A(B_m)-C(B_m)|, 
\end{equation}
where $M$ is the number of bins, $n$ is the total number of predictions, $B_m$ denotes the set of predictions falling into the $m$-th bin, and $A(B_m)$ and $C(B_m)$ represent the accuracy and average confidence in that bin, respectively.

Analogously, uncertainty calibration~\cite{uce} focuses on  aligning the predicted uncertainty with its actual  error, characterizing perfect calibration as
\begin{equation}
    \mathbb{P}(\hat{y} \neq y | H[\hat{p}]=u)=u,  \hspace{3mm} u\in [0,1]
    \label{eq:calibration_unc}
\end{equation}
where $H$ denotes the PE. This formulation implies that predictions with higher uncertainty (entropy) should fail more often in a predictable, probabilistic sense.
Analogous to ECE, the Uncertainty Calibration Error (UCE)~\cite{uce} quantifies the gap between prediction error and model uncertainty, i.e., 
\begin{equation}
    UCE = \sum_{m=1}^{M} \frac{|B_m|}{n} |E(B_m)-U(B_m)|, 
\end{equation}
 where $E(B_m)$ and $U(B_m)$ are the error and uncertainty of bin $m$.

The previous definitions were proposed for classification models. Hence, multiple works have extended the notion of model calibration for different tasks, such as regression. In regression settings, a calibrated model is expected to produce predictive distributions whose confidence 
intervals contain the true target values with the correct frequency.
Unlike classification, where calibration assesses the alignment between confidence and accuracy, calibration in regression focuses on the consistency between predicted uncertainty and the actual magnitude of errors. A well-calibrated regression model should assign higher uncertainty to inputs where prediction errors are likely to be large, and lower uncertainty where errors are expected to be small.
A common approach~\cite{calib_regression1,calib_regression2,uce2,ence} to express uncertainty in regression is through distributional regression, where the model outputs a full predictive distribution $P(y|x)$, such as a Gaussian parameterized by a predicted mean $\mu(x)$ and variance $\sigma^2(x)$. Under this formulation, a calibrated model satisfies the condition
\begin{equation}
    \mathbb{E}[(y - \mu(x))^2 |  \sigma^2(x)= \sigma^2] = \sigma^2,
    \label{eq:error_variance_matching}
\end{equation}
meaning that the squared error should match the predicted variance~\cite{calib_reg3, heteroscedastic2}. 

Overall, existing formulations of model calibration, whether in classification or regression, share a common principle: \textit{the model's predicted uncertainty should match its actual error.}  In classification, this manifests as the alignment between confidence and accuracy; in regression, it involves matching the predicted variance to the expected squared error or ensuring  that empirical coverage aligns with nominal intervals.

Further, to quantify miscalibration, various approaches rely on binning-based approximations. These methods discretize either predicted confidence or uncertainty scores into bins and compute the discrepancy between predicted and observed quantities within each bin. While effective, this strategy introduces design choices such as the number and placement of bins, which can significantly affect the sensitivity and resolution of the calibration estimate~\cite{ece,model_calibration}. Moreover, binning suffers from sample inefficiency in underpopulated regions and may hide finer-grained miscalibration patterns~\cite{bin1,ace}.

Building on these foundational formulations, we propose a novel method, called \name, for model calibration that is both generalizable and efficient, as detailed in Section~\ref{sec:main}.

\subsection{Related Work}
\label{sec:rw}


A significant body of work has explored methods to enhance calibration of NNs.
Early approaches leveraged Bayesian Neural Networks (BNNs)~\cite{BayesNN,BayesNN1,BayesNN2,BNN}, which place a prior distribution over model weights and infer a posterior to capture epistemic uncertainty~\cite{epis_alea_unc}. Despite their theoretical appeal, BNNs often suffer from significant computational overhead, making them difficult to scale to large models and datasets. To address this challenge, approximate Bayesian methods such as MC dropout~\cite{mcdropout1} and VI~\cite {Var_inf} have been proposed. While effective in certain settings, these approaches often trade off between computational tractability and predictive quality~\cite{unc_nn_eff}.
Notwithstanding, MC Dropout is the most popular method in the literature for uncertainty estimation of NNs~\cite{euat,uncertainty_survey,mcdropout1}. This method employs multiple dropout masks to simulate model uncertainty through multiple stochastic forward passes. To quantify models' uncertainty, the predictions are then aggregated using different statistical metrics such as PE, or variance.

An alternative line of research investigates deep ensembles~\cite{deep_ensembles}, where multiple models are independently trained to capture epistemic uncertainty. Ensembles have been shown to achieve strong empirical performance in both uncertainty quantification and calibration metrics, outperforming some Bayesian methods~\cite{ensemble_unc}. However, the cost of training and maintaining multiple networks becomes prohibitive for large-scale architectures such as language models or vision transformers.
Furthermore, modeling techniques that inherently produce uncertainty estimates for each prediction, such as Deep Gaussian Processes~\cite{deepGP} or Laplace approximations~\cite{laplace}, can be leveraged for uncertainty estimation. However, these approaches constrain the flexibility of architectural design~\cite{uncertainty_survey}.

Alongside uncertainty quantification, Guo et al.~\cite{model_calibration} have emphasized the importance of NNs calibration by demonstrating that  NNs often suffer from overconfidence.
Thus, parallel to advances in uncertainty modeling, post-hoc calibration methods (e.g., Platt Scaling~\cite{platt}, Isotonic Regression~\cite{isotonic_regression}, Temperature Scaling~\cite{model_calibration}, or Beta Calibration~\cite{beta_calib}) have been developed to adjust model confidences after training. 
While these post-hoc calibration techniques are computationally efficient and widely adopted, they primarily adjust model confidence without improving the underlying uncertainty estimation. As a result, their performance is often sensitive to both model architecture and dataset~\cite{cals}, and degrade under distributional shifts, leading to sub-optimal reliability in OOD scenarios~\cite{calib_ood,model_calibration}.

Beyond post-hoc calibration, a prominent line of work focuses on incorporating uncertainty estimation directly into the training process. These uncertainty-aware training methods jointly optimize for accuracy and uncertainty calibration during model learning. To this end, several loss functions have been proposed that augment standard objectives with uncertainty terms. For instance, Focal Loss~\cite{focal_loss} and Label Smoothing~\cite{focal_loss1} introduce mechanisms to reduce overconfidence, thereby implicitly improving calibration by softening target distributions or down-weighting easy examples.
Building on this idea, Krishnan et al.~\cite{acc_unc_calib} introduced the Accuracy versus Uncertainty Calibration (AvUC) loss, which jointly optimizes for both predictive performance and well-calibrated uncertainty estimates, and integrates temperature scaling with AvUC. Building on this, Karandikar et al.~\cite{soft_calibration} proposed soft calibration error metrics as a continuous relaxation of traditional binning-based estimators, and Gupta et al.~\cite{bin_free_calib} introduced a binning-free calibration approach, circumventing the discretization inherent in traditional calibration metrics.
Complementarily, Einbinder et al.~\cite{Uncertainty_loss1} introduced an uncertainty-aware conformal loss, which integrates a conformal prediction-based uncertainty term to better align confidence intervals with empirical outcomes. Further, Shamsi et al~\cite{ce_pe} added a new term to account for the model's uncertainty that can be determined through the PE or the ECE. More recently, CALS~\cite{cals} incorporates class-wise uncertainty to dynamically weighted the terms in the loss function placing greater emphasis on uncertain predictions, and EUAT~\cite{euat} employs two distinct loss functions based on whether the inputs are correctly or incorrectly predicted by the model in order to ensure high uncertainty for mispredictions and low uncertainty for correct ones while preserving model quality. However, these approaches only work for classification models.

Model calibration has been also studied across various domains, including regression and language modeling. In the regression setting, Quantile-Calibrated Regression\cite{calib_regression2} and Distribution-Calibrated Regression\cite{calib_regression2} introduce calibration techniques based on confidence intervals and distributional alignment, respectively.
An orthogonal line of research focuses on modeling heteroscedastic uncertainty~\cite{heteroscedastic1,heteroscedastic2}, where models are trained to predict both the mean and variance, enabling them to capture varying noise levels across data points.

In the context of language models, recent work has examined uncertainty in generative and structured prediction tasks such as machine translation~\cite{unc_translation}, question answering~\cite{calib_transformers}, and open-ended language generation~\cite{unc_language}. Approaches for uncertainty estimation in language models are commonly categorized as either black-box\cite{blacl_unc_lang} or white-box, with the latter further divided into information-based methods\cite{unc_language1,unc_language2,unc_language3,unc_language4}, ensemble-based approaches\cite{ens_unc_lang}, and density-based techniques\cite{dens_unc_lang1,dens_unc_lang2,dens_unc_lang3,dens_unc_lang4}.
Recently, Fadeeva et al.~\cite{unc_lang} introduced a unified framework implementing several state-of-the-art uncertainty estimation techniques for large language models (LLMs) in text generation tasks, offering standardized interfaces. 
These methods collectively demonstrate the potential of uncertainty-aware objectives to enhance model trustworthiness. However, many remain tailored to specific architectures or tasks, motivating the need for scalable, general-purpose uncertainty-aware training frameworks.

In contrast to existing work, \name  aims to achieve efficient and task-agnostic uncertainty estimation, suitable for both classification and regression problems, as well as language models. Our approach emphasizes preserving high predictive performance while providing calibrated uncertainty estimates without the computational burden typically associated with Bayesian or ensemble-based methods.

\section{\nameAll}
\label{sec:main}

\name is a task-agnostic method and, consequently, we adopt a general definition of model calibration that can be applied, e.g., both to regression and classification. 
Let a model $f$: $\mathcal{X} \rightarrow \mathcal{Y}$ produce both a prediction \( \hat{y} = f(x) \) and an associated uncertainty estimate \( u(\hat{y}) \in \mathbb{R}_{\ge 0} \) for an input \( x \in \mathcal{X} \), and consider a task-specific error function
$\mathcal{E}(y, \hat{y})$: $\mathcal{Y} \times \mathcal{Y} \rightarrow \mathbb{R}_{\ge 0}$.
A model is perfectly calibrated if, for all uncertainty levels $u$,
\begin{equation}
    \mathbb{E}[\mathcal{E}(y, \hat{y}) \mid u(\hat{y}) = u] = u.
    \label{eq:calibration_general}
\end{equation}
This expresses the principle that the model's predicted uncertainty should match its expected error (see Section~\ref{sec:problem}). This generic formulation can be adapted to the settings of classification and regression by specializing the definition of $u(x)$ and $\mathcal{E}(y, \hat{y})$.


%

For classification, the error function and uncertainty estimation can be defined, respectively, as $\mathcal{E}(y, \hat{y}) = \mathbbm{1}\{ y \neq \hat{y} \}$ and $u(x)=1-\hat{p}(x)$, where $\hat{p}(x)=\max_y P(\hat{y} = y | x)$ represents the model's confidence. Adopting these definitions, Equation~\ref{eq:calibration_general} reduces to the condition $\mathbb{P}(\hat{y} \neq y | 1-\hat{p}=u)=u$, which is equivalent to the confidence-based calibration for classification defined in Equation~\ref{eq:calibration}.

For regression problems, the error function can be defined as the squared error $\mathcal{E}(y, \hat{y}) = (y - \hat{y})^2$, and the model's uncertainty can be quantified via the predicted variance, $u(x) = \sigma^2(x)$.  In this case, calibration corresponds to the condition
$\mathbb{E}[(y - \hat{y})^2 \mid \sigma^2(x) = u] = u$, 
which corresponds to the formulation of a calibrated model in regression settings (see Equation~\ref{eq:error_variance_matching}).

Based on this general formulation of the calibration problem, we aim at solving an optimization problem whose goal is to minimize the discrepancy between uncertainty estimates and predictive error.
Specifically, we operationalize this principle by leveraging Equation~\ref{eq:calibration_general}, which captures the desired agreement between uncertainty and predictive performance at the individual input level. 
\name eschews resorting to coarse-grained binning or full distributional comparisons and adopts a lightweight and pragmatic approach that leverages the task-specific loss function $L_{e}$ (e.g., Cross-entropy (CE) for classification or Mean Squared Error (MSE) for regression) as a differentiable proxy for the instance-level error function $\mathcal{E}(y, \hat{y})$. 

More in detail, \name incorporates the calibration formulation directly into the training objective to jointly optimize for both predictive quality and uncertainty calibration. To this end, \name complements the standard task-specific loss function, $L_{e}$, using during conventional training, with a second loss term, serving as calibration regularizer, which is defined as the squared difference between   $L_{e}$ and the predicted uncertainty $u$.
\begin{equation}
    L(y,\hat{y}) =  \alpha \cdot L_{e}(y,\hat{y}) + (1-\alpha) \cdot \left( L_{e}(y,\hat{y}) - u(\hat{y}) \right)^2,
     \label{eq:loss}
\end{equation}
where 
$u$ represents the uncertainty metric, and $\alpha \in [0, 1]$ weights the trade-off between error and calibration.
The squared divergence term ensures that the loss remains continuous and differentiable, making it compatible with gradient-based optimization. 

To enhance robustness, \name estimates both predictions and uncertainties during training and inference by employing stochastic inference techniques such as MC Dropout, which aggregates outputs from multiple forward passes through the model. 
This enables both the estimation of predictive uncertainty and an improvement in prediction quality. 

To summarize, we formulate model calibration as minimizing the difference between the expected loss of a prediction and its associated uncertainty. To achieve this, we propose a loss function that jointly optimizes for both predictive error and calibration quality. Our approach leverages loss-based proxies to estimate model error, enabling direct comparison with uncertainty estimates. This eliminates the need for costly distributional comparisons or binning-based approximations, resulting in a fully differentiable and scalable method, compatible with modern neural architectures, facilitating the training of models that are both accurate and uncertainty-aware.

\section{Evaluation}
\label{sec:eval}

This section is devoted to evaluating \name, across a diverse set of tasks to demonstrate its effectiveness in aligning predicted uncertainty with model error, while preserving high predictive performance. Our experiments aim to answer the following key questions:

\textcolor{white}{n}\hspace{1mm}$\bullet$ \textit{\textbf{Calibration Quality:}} Does \name improve the alignment between uncertainty estimates and observed errors compared to existing approaches? \\
\textcolor{white}{n}\hspace{1mm}$\bullet$ \textit{\textbf{Predictive Performance:}} Does incorporating calibration objectives degrade task performance? \\
\textcolor{white}{n}\hspace{1mm}$\bullet$ \textit{\textbf{Generality:}} How well does \name perform across diverse tasks, model architectures, and uncertainty metrics? \\
\textcolor{white}{n}\hspace{1mm}$\bullet$ \textit{\textbf{Efficiency:}} Is \name computationally efficient?


\subsection{Experimental Setup, Benchmarks, and Baselines}
\label{sec:expe}

\textbf{Datasets and Models.}
Our evaluation spans a diverse set of tasks across vision, regression, and language domains. For image classification, we use four model–dataset pairs: ResNet-50~\cite{resnet} on ImageNet~\cite{imagenet}, Wide-ResNet-28x10~\cite{wide_resnet} on CIFAR-100~\cite{cifar10}, ResNet-18 on CIFAR-10, and ResNet-18 on SVHN~\cite{svhn}.
For regression, we consider three ResNet models for tabular data~\cite{resnet_reg} trained on the New York City Taxi Fare Prediction (NYC Taxi)~\cite{nyc_taxi}, Million Song (MSD)~\cite{music}, and Boston Housing~\cite{boston} datasets. In the language domain, we evaluate the T5 encoder-decoder transformer~\cite{t5} on two tasks: summarization using the XSum dataset~\cite{xsum} and English-to-French machine translation using the OpusBooks dataset~\cite{opus_book}.
We further include a binary classification task using ResNet-18 on CIFAR-10 (predicting whether an image contains a cat), and an out-of-distribution detection task based on distributional shifts introduced via corrupted inputs. 
Additional details to ensure reproducibility are provided in the Technical Appendix.

\textbf{Baselines.}
For classification, we compare \name against the standard CE loss, model calibration with Isotonic regression (Iso. Reg.)~\cite{isotonic_regression}, DEUP~\cite{deup}, deep ensembles with five learners~\cite{deep_ensembles}, CALS~\cite{cals}, and a combined CE with Predictive Entropy (CE+PE) loss~\cite{ce_pe}.
In regression, we benchmark against models trained with MSE, Negative Log-Likelihood (NLL), MSE with PE~\cite{ce_pe}, quantile calibration (Q-Reg)~\cite{calib_regression1}, DEUP, and deep ensembles.
For language models, we compare our approach against standard CE-trained models using multiple uncertainty estimation methods: maximum sequence probability~\cite{unc_lang}, perplexity~\cite{unc_language4}, mean token entropy~\cite{unc_language4}, Monte Carlo sequence entropy~\cite{unc_language3}, pointwise mutual information (PMI)~\cite{unc_language2}, and conditional PMI~\cite{unc_language3}.
To calibrate post-training methods and train DEUP’s auxiliary error predictor, we reserve 10\% of the data as a validation set. We apply isotonic regression for post-hoc calibration, which, in our experiments,  outperforms alternative approaches such as Platt scaling~\cite{platt}, temperature scaling~\cite{model_calibration}, and beta calibration~\cite{beta_calib}. For fairness, we ensure that all methods relying on a validation set use the same set consistently.

\textbf{Evaluation Metrics.}
We evaluate all models using a comprehensive set of metrics tailored to classification, regression, and language modeling tasks.
In classification, we report seven metrics: (i) classification error rate, (ii) ECE, (iii) Uncertainty Accuracy (uA)~\cite{euat}, and (iv) Uncertainty Area Under the Curve (uAUC)~\cite{euat}, both derived from the Uncertainty Confusion Matrix~\cite{unc_conf_matrix}, (v) Pearson correlation between prediction residuals and estimated uncertainties (Corr. w/ res.)~\cite{deup}, (vi) Wasserstein distance between the uncertainty distributions of correct and incorrect predictions~\cite{Wasserstein_distance}, and (vii) average training time per epoch.
Unless otherwise specified, all uncertainty estimates are computed via MC Dropout using normalized PE. For DEUP, which estimates uncertainty from loss values via a meta-model, we normalize the loss values to ensure fair comparison.
In regression, we use the following metrics: (i) MSE, (ii) ECE, (iii) correlation between MSE and uncertainty estimates (via PE or variance), (iv) NLL, (v) Expected Normalized Calibration Error (ENCE)~\cite{ence}, and (vi) Area Under the Sparsification Error curve (AUSE)~\cite{unc_reg_metric}.
For language models, we adopt the Prediction Rejection Ratio (PRR) metric, as recommended by Fadeeva et al.~\cite{unc_lang}, using ROUGE-2 for summarization and BLEU for translation to quantify task-specific quality. 

\subsection{Experimental results}
Next, we report the results obtained using \name across the different domains evaluated. Additional results evaluating the efficiency of \name are presented in the Appendix. 

\setlength{\tabcolsep}{4pt}
\begin{table*}[t]
    \caption{Comparison of \name against the baselines on classification models.}
    \label{tab:classification}
    \centering
    \begin{tabular}{clccccccc}
        \toprule
        \multirow{2}{*}{\textbf{Benchmark}} & \multirow{2}{*}{\textbf{Baseline}} & \multirow{2}{*}{\textbf{Error}$\downarrow$} & \multirow{2}{*}{\textbf{ECE}$\downarrow$} & \multirow{2}{*}{\textbf{uA}$\uparrow$} &  \multirow{2}{*}{\textbf{uAUC}$\uparrow$} & \textbf{Corr.} & \textbf{Wasser.} & \textbf{Time per} \\  
        & & & & & &\textbf{w/ res.}$\uparrow$ & \textbf{dist.}$\uparrow$ &\textbf{epoch}[s]$\downarrow$\\[2mm]  \hline
        %
        & \textbf{\name} &	0.47 & \textbf{0.22} & \textbf{0.82} & \textbf{0.89} & \textbf{0.69} & 0.26  & 7239 \\
        & \textbf{EUAT}         & \textbf{0.44} & \textbf{0.22}  & 0.8 & 0.88 & 0.66 & \textbf{0.3}  & 9882 \\
        ResNet50 & \textbf{CE}           &	0.51 & \textbf{0.22} & 0.75 & 0.81 & 0.55 & 0.22   & \textbf{3844}\\
        & \textbf{Iso. Reg.} &	0.55 & 0.27 & 0.74 & 0.81 & 0.54 & 0.19   & 3898\\
        ImageNet & \textbf{DEUP}         &	0.52 & 0.43 & 0.54 & 0.59 & 0.2 & 0.01   & 3914\\
        & \textbf{Ensemble}     &	0.51 & 0.25 & 0.75 & 0.81 & 0.54 & 0.21   & 7355\\
        & \textbf{CALS}         &	0.53 & 0.24 & 0.75 & 0.81 & 0.54 & 0.21   & 3913\\
        & \textbf{CE+PE}        &	0.52 & 0.38 & 0.76 & 0.79 & 0.53 & 0.14   & 9824\\
        \hline
        & \textbf{\name}   & \textbf{0.26}  & 0.15  & \textbf{0.85} & 0.87 &	\textbf{0.72}	& 0.19& 750 \\
        & \textbf{EUAT}   &	0.27  & 0.16 & \textbf{0.86} &	\textbf{0.89} &0.71	&\textbf{ 0.22}& 1214\\
       Wide-ResNet & \textbf{CE}           &	0.3 & 0.24 & 0.79 &	0.77 &	0.55	& 0.13  & \textbf{641} \\
        & \textbf{Iso. Reg.} &	0.31 & 0.15 & 0.79 &	0.84 &	0.58	& 0.23  & 655 \\
       Cifar100 & \textbf{DEUP}         &	0.33 & 0.25 & 0.68 &	0.56 &	0.19	& 0.03  & 661\\
        & \textbf{Ensemble}     &	0.33 & 0.29 & 0.74 &	0.7 &	0.47	& 0.09  & 1407 \\
        & \textbf{CALS}        &	\textbf{0.26} & \textbf{0.11} & 0.81 &	0.83 &	0.59	& 0.23  & 653\\
        & \textbf{CE+PE}       &	0.3 & 0.25 & 0.78 &	0.74 &	0.52	& 0.1  & 980 \\
        \hline
        & \textbf{\name} &	0.1 &	\textbf{0.01}  & \textbf{0.92} &\textbf{0.93} & \textbf{0.65} & 0.34 & 235\\
        & \textbf{EUAT}  &	0.1  & 0.02  & 0.91 &	0.92 & 0.63 &	\textbf{0.41}& 367 \\
        ResNet18 & \textbf{CE}          & 0.1 & 0.03 & 0.91 & 0.87 & 0.58 & 0.27  & \textbf{156} \\
        & \textbf{Iso. Reg.} & 0.11 & 0.03 & 0.9 & 0.89 & 0.53 & 0.33  & 174\\
        Cifar10 & \textbf{DEUP}        & \textbf{0.09} & 0.04 & \textbf{0.92} & 0.54 & 0.3 & 0.03  & 177 \\
        & \textbf{Ensemble}     & \textbf{0.09}  & 0.04 & \textbf{0.92} & 0.84 & 0.55 & 0.22 & 294\\
        & \textbf{CALS}         & 0.1 & \textbf{0.01}  & 0.91 & 0.88 & 0.56 & 0.28 & 159\\
        & \textbf{CE+PE}      & 0.1 & 0.05  & 0.91 & 0.84 & 0.57 & 0.21 & 284 \\
        \hline 
        & \textbf{\name} &\textbf{0.04} & \textbf{0.01} & \textbf{0.96} & 0.89	& \textbf{0.65} & 0.39  & 419 \\
        & \textbf{EUAT} & 0.05 & \textbf{0.01} & \textbf{0.96} &	\textbf{0.93}	& 0.64 &	\textbf{0.48}   & 5377 \\
        ResNet18& \textbf{CE}           & 0.05 & 0.02 & 0.95 &	0.84	& 0.57 &	0.23 & \textbf{344}\\
        & \textbf{Calibration} & 0.05 & 0.03 & 0.95 &	0.9	& 0.54 &	0.35 & 353\\
        SVHN& \textbf{DEUP}         & \textbf{0.04} & 0.02 & \textbf{0.96} &	0.56	& 0.31 &	0.04 & 356\\
        & \textbf{Ensemble}     &\textbf{0.04} & 0.03 & \textbf{0.96} &	0.76	& 0.52 &	0.16 & 715 \\
        & \textbf{CALS}         & \textbf{0.04} & \textbf{0.01}  & \textbf{0.96} &	0.87	& 0.57 &	0.26  & 364\\
        & \textbf{CE+PE}        & 0.05 & 0.03 & \textbf{0.96} &	0.8	& 0.55 &	0.18 &  548\\
        \bottomrule
    \end{tabular}
\end{table*}
\raggedbottom

\textbf{Image Recognition Models: }
We begin by presenting the results for four image recognition tasks in Table~\ref{tab:classification}, using the following model–dataset pairs: ResNet-50 on ImageNet, Wide-ResNet-28x10 on CIFAR-100, ResNet-18 on CIFAR-10, and ResNet-18 on SVHN. 
\name achieves consistently strong performance across all benchmarks, outperforming or matching state-of-the-art methods in both predictive accuracy and uncertainty estimation, achieving the best performance in 15 of 24 comparisons. Further, in the few cases (e.g., using the Wasserstein distance) where alternative methods perform comparably to \name, the differences are minor. 
On ResNet50/ImageNet, \name attains the highest uA (0.82) and uAUC (0.89), with competitive calibration (ECE = 0.22). For CIFAR100, it yields the lowest error (0.26) and highest residual correlation (0.72), indicating precise uncertainty alignment with model errors.
On ResNet18 with CIFAR10 and SVHN, \name leads in calibration error (ECE=0.01) and maintains top-tier performance in uncertainty metrics (uA=0.92/0.96, uAUC=0.93/0.89, respectively), while remaining computationally efficient. Notably, \name outperforms DEUP and EUAT in several settings without sacrificing runtime or scalability.

\name consistently achieves the highest correlation with residuals across benchmarks, which is expected given that it is explicitly designed to align the uncertainty with prediction errors. This strong alignment not only validates \name’s uncertainty estimates but also contributes to its competitive calibration performance, as reflected in its consistently low ECE scores. This dual strength highlights \name's ability to deliver both reliable confidence measures and practical predictive uncertainty.

\setlength{\tabcolsep}{4pt}
\begin{table*}[t]
    \caption{Comparison of \name against the baselines on regression problems.}
    \label{tab:regression}
    \centering
    \begin{tabular}{clcccccccc}
        \toprule
         \multirow{3}{*}{\textbf{Bench.}} &  \multirow{3}{*}{\textbf{Baseline}} &  \multirow{3}{*}{\textbf{MSE}$\downarrow$} &  \multirow{3}{*}{\textbf{ECE}$\downarrow$} & \textbf{Corr. } & \textbf{Corr. } &  \multirow{3}{*}{\textbf{NLL}$\downarrow$}  &  \multirow{3}{*}{\textbf{ENCE}$\downarrow$} &  \multirow{3}{*}{\textbf{AUSE}$\downarrow$} & \textbf{Time } \\
        &  & & & \textbf{Error} & \textbf{Error} & & &  &\textbf{per}\\
        &  & & & \textbf{PE}$\uparrow$ & \textbf{Var}$\uparrow$ & & &  &\textbf{epoch}[s]$\downarrow$ \\[2mm] 
        \hline 
        & \textbf{\name} & \textbf{16.9} 	& \textbf{0.02} & \textbf{0.30} & \textbf{0.39} & \textbf{2.33}	& \textbf{0.82} & 0.03 & 6221 \\
        & \textbf{MSE} & 18.4	& 0.28 & 0.28 & 0.22 & 11.6	& 3.66 & 0.08 & \textbf{1708} \\
         NYC& \textbf{Q-Reg}  & 18.4	& 0.49 & 0.22 & 0.29 & 69.9	& 4.45 & 0.27 & 1767 \\
        & \textbf{DEUP} & 19.8	& 0.39 & 0.28 & 0.19 & \textbf{2.32} & 0.81 & 0.04 & 1864  \\
        taxi& \textbf{Ensemble} & 17.5 & 0.29 & \textbf{0.30} & 0.32 & 10.2	& 3.43 & 0.08 & 3590 \\
        & \textbf{NLL} & 37.0 	& 0.10 & 0.03 & 0.01 & 3.75	& 1.36 & \textbf{0.02} & 1973 \\
        & \textbf{MSE+PE} & 18.4& 0.67 & \textbf{0.30} & 0.25 & 307 & 14.8 & 0.21 & 7319 \\
        \hline
         \multirow{7}{*}{MSD}& \textbf{\name} & \textbf{112.9} & \textbf{0.09} & \textbf{0.15} & \textbf{0.1} & \textbf{3.77}  & \textbf{0.96} & \textbf{0.01} & 253\\
        & \textbf{MSE} & 130.5 & 0.34 & 0.12 &\textbf{ 0.1} & 37.9 & 7.54 & 0.04 & \textbf{35}\\
        & \textbf{Q-Reg} & 132.7 & 0.40 & -0.01 & -0.22 & 63.9 & 8.71 & 0.32 & \textbf{35}\\
        & \textbf{DEUP} & 134.4 & 0.42 & 0.14 & -0.01 & 3.94 & 1.83 & \textbf{0.01}& 38\\
        & \textbf{Ensemble} & 142.4 & 0.42 & 0.01 & 0.03 & 47.5 & 8.44 & 0.03 & 71\\
        & \textbf{NLL} & 134.8 & 0.11  & 0.09 & 0.05 & 8.19 & 2.76 & \textbf{0.01} & 44\\
        & \textbf{MSE+PE} & 137.3 & 0.43 & 0.02 & 0.01 & 48.3 & 8.61 & 0.03 & 268\\
        \hline
        & \textbf{\name} & 15.3 & 0.02 & \textbf{0.42} & \textbf{0.57} & \textbf{2.08} & \textbf{0.68} & 0.03 & 0.9 \\
        & \textbf{MSE} & \textbf{14.3} & 0.03 & 0.41 & 0.56 & 4.15 & 1.63 & 0.5 & \textbf{0.12} \\
        Boston & \textbf{Q-Reg} & 18.6 & 0.1 & 0.39 & 0.36 & 59.7 & 2.26	 & 0.12 & 0.13\\
        & \textbf{DEUP} & 22.7 & 0.38  & 0.06 & -0.05 & 14.5 & 3.77 &\textbf{0.01} & 0.13 \\     
        Housing & \textbf{Ensemble} & 15.2 & 0.05 & 0.37 & 0.46 & 3.48	& 1.4	& 0.05	& 0.28\\
        & \textbf{NLL} & 18.1 	& \textbf{0.01} & -0.01 & 0.08 & 3.08 & 0.95 & \textbf{0.01} & 0.16 \\
        & \textbf{MSE+PE} &  16.6 & 0.06 & 0.36 &	0.57 &	5.84 & 2.21 &	0.07 &	0.96 \\
        \bottomrule
    \end{tabular}
\end{table*}
\raggedbottom

\textbf{Regression Models: }
In regression benchmarks (see Table~\ref{tab:regression}), \name achieves consistently strong performance in both predictive accuracy and uncertainty estimation. On NYC Taxi, \name delivers the lowest ECE and ENCE (0.02 and 0.82) and best overall calibration, while achieving competitive MSE (16.9) and highest correlation between error and PE (0.30) or variance (0.39). On MSD, \name also attains the lowest ECE and ENCE  (0.09 and 0.96), and best AUSE and correlation between error and uncertainty among all methods, alongside a favorable trade-off in log-likelihood and runtime. For Boston Housing, \name shows a strong correlation between error and PE (0.42) or variance (0.57), while maintaining competitive NLL and the lowest ECE and ENCE (0.02 and 0.68). These results further support that \name not only models uncertainty accurately but does so with minimal calibration error, outperforming other methods, especially in metrics it is tailored to optimize
To sum up,  \name demonstrates strong performance across diverse calibration metrics, including ECE, ENCE, and the correlation between error and uncertainty. Remarkably, it also achieves the best (or near-best) NLL, even without being explicitly trained to optimize it.

\begin{table*}[t]
    \caption{Comparison of \name using PRR$\uparrow$ against the baselines on language models.}
    \label{tab:nlp}
    \centering
    \begin{tabular}{l|cc|cc}
        \toprule
        \multirow{2}{*}{\textbf{Uncertainty Method}} & \multicolumn{2}{c|}{\textbf{Summarization}} & \multicolumn{2}{c}{\textbf{Translation}}\\
        &  \textbf{\name} & \textbf{CE}& \textbf{\name} & \textbf{CE} \\  \hline 
        Quality Metric (ROUGE-2/BLEU) & \textbf{4.75e-2} & 4.54e-2 & \textbf{0.242} & 0.206\\
        Maximum Sequence Probability & \textbf{6.69e-2} & 6.25e-2 & \textbf{0.304} & 0.262\\
        Perplexity& \textbf{6.61e-2} & 6.22e-2  & \textbf{0.309} & 0.262\\
        Mean Token Entropy & \textbf{6.57e-2}  & 6.09e-2 & \textbf{0.307} & 0.260 \\
        Monte Carlo Sequence Entropy & \textbf{7.72e-2} & 7.01e-2  & \textbf{0.294} & 0.252 \\
        Monte Carlo Normalized Sequence Entropy & \textbf{7.68e-2}  & 7.11e-2  & \textbf{0.302} & 0.257\\
        Pointwise Mutual Information & \textbf{6.75e-2} & 6.41e-2 & \textbf{0.246}& 0.213 \\
        Conditional Pointwise Mutual Information & \textbf{6.74e-2} & 6.42e-2 & \textbf{0.227} & 0.198 \\
        Time per epoch [s] & 60573 & \textbf{60257} &29658& \textbf{29051}\\
        \bottomrule
    \end{tabular}
    
\end{table*}
\raggedbottom

\textbf{Language Models: }
Table~\ref{tab:nlp} presents the comparison of the \name framework against the CE baseline across summarization and translation tasks, emphasizing performance and computational efficiency. Across all uncertainty methods, \name consistently outperforms CE in both ROUGE-2 and BLEU metrics, indicating enhanced quality of generated outputs. On average, for summarization and translation tasks, \name improves by 7\% and 17\% the results obtained using CE, respectively. Notably, \name demonstrates better uncertainty estimations while maintaining comparable training efficiency, as evidenced by similar time per epoch across tasks. Note that, to minimize the overhead introduced by \name, the PE for each prediction is computed using a single forward pass.

\begin{table*}[t]
    \caption{Comparison of \name against the baselines on out-of-distribution detection task.}
    \label{tab:ood}
    \centering
    \begin{tabular}{lccccccc}
        \toprule
        \multirow{2}{*}{\textbf{Baseline}} & \multirow{2}{*}{\textbf{Error}$\downarrow$} & \multirow{2}{*}{\textbf{ECE}$\downarrow$}  & \multirow{2}{*}{\textbf{uA}$\uparrow$} &  \multirow{2}{*}{\textbf{uAUC}$\uparrow$} & \textbf{Corr.} & \textbf{Wasser.} & \textbf{Time per} \\
        &  & & & & \textbf{w/ res.}$\uparrow$ & \textbf{dist.}$\uparrow$ &\textbf{epoch}[s]$\downarrow$\\[2mm]  \hline 
        \textbf{\name}  & 0.539 & \textbf{0.142}  &\textbf{ 0.758} & 0.752 & \textbf{0.559} & 0.191 & 731\\
        \textbf{EUAT}  & 0.489 & \textbf{0.143}  & \textbf{0.754} & \textbf{0.796} & 0.529 & \textbf{0.255} & 820\\
        \textbf{CE}   & 0.539  & 0.292  &  0.691 & 0.676 & 0.311 & 0.126& \textbf{636}\\
        \textbf{Iso. Reg.} & 0.619  & 0.497  & 0.619 & 0.553 & 0.110 & 0.036 & 638\\
        \textbf{DEUP}    & 0.617 & 0.551  & 0.617 & 0.509 & 0.044 & 0.003 & 639\\
        \textbf{Ensemble} & 0.555 & 0.237 & 0.753 & 0.734 & 0.426 & 0.177 & 1259\\
        \textbf{CALS}  & \textbf{0.464} & 0.216 & 0.686 & 0.716 & 0.383 & 0.147  & 662\\
        \textbf{CE+PE} & 0.568   & 0.418 & 0.681 & 0.663 & 0.299 & 0.103 & 731\\
        \bottomrule
    \end{tabular}
\end{table*}
\raggedbottom

\textbf{Out-Of-Distribution Detection Task: }
Next, on Table~\ref{tab:ood}, we evaluate the effectiveness of \name to detect OOD examples based on the predicted uncertainty. 
To that end, we trained a ResNet18 using the Cifar10 dataset and then tested it using a corrupted version with Gaussian noise of the same dataset (called Cifar10-C~\cite{cifar10-c}). 
For this task, \name achieves competitive performance across all key metrics. It ranks among the best in ECE (0.142) and leads in correlation with residuals (0.559), indicating highly calibrated and reliable uncertainty estimates. While EUAT achieves the best uAUC (0.796) and Wasserstein distance (0.255), \name still maintains similar scores in both (uAUC=0.752, Wasserstein=0.191) with a lower computational cost. Overall, \name offers a strong trade-off between accuracy, calibration, and efficiency for detecting OOD samples.

\setlength{\tabcolsep}{1.5pt}
\begin{table*}[t]
    \caption{Comparison of \name against the baselines on a binary classification problem.}
    \label{tab:binaryCifar10}
    \centering
    \begin{tabular}{lcccccccccccc}
        \toprule
        \multirow{3}{*}{\textbf{Baseline}} & \textbf{Error} & \textbf{Error}	& \multirow{3}{*}{\textbf{F1}$\uparrow$} & \multirow{3}{*}{\textbf{Prec.}$\uparrow$} & \multirow{3}{*}{\textbf{TPR}$\uparrow$} & \multirow{3}{*}{\textbf{TNR}$\uparrow$} & \multirow{3}{*}{\textbf{ECE}$\downarrow$} & \multirow{3}{*}{\textbf{uA}$\uparrow$} &  \multirow{3}{*}{\textbf{uAUC}$\uparrow$} & \textbf{Corr.} & \multirow{3}{*}{\shortstack{\textbf{Wasser.}\\ \textbf{dist.}$\uparrow$}}   & \textbf{Time} \\
        & \textbf{w/o} & \textbf{w/} & & & &  & & & & \textbf{w/} & &\textbf{per} \\
        & \textbf{flip}$\downarrow$ & \textbf{flip}$\downarrow$ & & & &  & & & & \textbf{res.}$\uparrow$ & &\textbf{epoch}[s]$\downarrow$ \\ \hline        
        \textbf{\name} & 0.17 & \textbf{0.13} & \textbf{0.86} &\textbf{0.89} & \textbf{0.87} & \textbf{0.91} & \textbf{0.11} & 0.84	& 0.75	& \textbf{0.46} & 0.36 & 45\\
        \textbf{EUAT}  & 0.15 & 0.14 & \textbf{0.86} & 0.86 & 0.86 & 0.86 & \textbf{0.11} & \textbf{0.86} & \textbf{0.82} & 0.45 & \textbf{0.41} & \textbf{203}\\
        \textbf{CE}   & 0.15 & 0.22 & 0.76 & 0.82 & 0.71 & 0.85 &  0.12 &  0.85 & 0.76 & 0.44 & 0.35 & 30\\
        \textbf{Iso. Reg.} & 0.16 & 0.29 & 0.73 & 0.67 & 0.81 & 0.60 & 0.31 & 0.78 & 0.78 & 0.37 & 0.36 & 31 \\
        \textbf{DEUP}& 0.17 & 0.18 & 0.81 & 0.86 & 0.77 & 0.87 & 0.12 & 0.83 & 0.53 & 0.21 & 0.02 & 31 \\
        \textbf{Ensemble} & \textbf{0.15} & 0.16 & 0.83 & 0.84 & 0.82 & 0.85& \textbf{0.11} & \textbf{0.86} & 0.68 & 0.36 & 0.22 &59 \\
        \textbf{CALS} & 0.16 & 0.16 & 0.83 & 0.85 & 0.82 & 0.85 & 0.12  & 0.83 & 0.75 & 0.41 & 0.33 & 31\\
        \textbf{CE+PE} & 0.18 & 0.23 & 0.76 & 0.82 & 0.71 & 0.84 & 0.15 & 0.81 & 0.69 & 0.36 & 0.27 & 45 \\
        \bottomrule
    \end{tabular}
\end{table*}
\raggedbottom

\textbf{Binary Classification Problem: }
On the binary classification task, \name achieves balanced performance across accuracy, calibration, and uncertainty metrics. It matches the best F1 score (0.86) and outperforms all baselines in precision (0.89), while maintaining strong calibration (ECE=0.11) and a high correlation with residuals (0.46). Although EUAT achieves slightly better uA, uAUC, and Wasserstein distance (probably due to its design goal that aims to separate the set of wrong and correct predictions), \name demonstrates more consistent performance across a wider range of metrics and offers better computational efficiency (the training time reduces by 4.5$\times$). Overall, \name delivers reliable and well-calibrated predictions while preserving competitive accuracy, confirming its effectiveness for uncertainty-aware binary classification.

\section{Conclusion}
\label{sec:conclusion}

We introduced  \name, a general calibration framework for NNs that aligns predictive uncertainty with model error through a scalable and differentiable loss. Unlike traditional approaches relying on binning or distributional comparisons, our method operates directly on summary statistics and loss-based proxies, making it broadly applicable to classification, regression, and language modeling tasks.
Our experiments across diverse benchmarks demonstrate that our method enhances uncertainty estimation while preserving predictive accuracy and introducing only minimal computational overhead.
Overall, our method provides a general and efficient approach to uncertainty calibration that is compatible with modern deep learning architectures and scalable to real-world applications.


\newpage
\bibliographystyle{plain}
\bibliography{mybibfile}



\appendix
\section*{Technical Appendix}
\section{Experimental Setup and Implementation Details}
\label{app:a}

Across all experiments, models include dropout layers with a fixed dropout rate of 0.3 to enable stochastic inference for uncertainty estimation. For classification tasks, we train all models using stochastic gradient descent (SGD) with momentum 0.9 and a batch size of 64. The learning rate and weight decay are set to 0.01 and $10^{-5}$ for ResNet-50 on ImageNet, and to 0.1 and 0 for all other classification models. These models are trained for 60 epochs, except for the binary classification task, which is trained for 200 epochs. In these models, increasing the number of MC dropout samples has minimal impact on the quality of uncertainty estimation. Therefore, we use only 5 forward passes in our experiments to balance computational efficiency and performance.

For regression tasks, we employ SGD with a learning rate of $10^{-4}$, momentum of 0.9, and a batch size of 256 for both the NYC Taxi and MSD datasets, training each model for 30 epochs. For the Boston Housing dataset, we use a higher learning rate of $10^{-3}$ and a smaller batch size of 64, training for 200 epochs. 
In this case, the number of MC dropout samples significantly affects the quality of uncertainty estimation (see Appendix~\ref{app:b}). Consequently, we increase the number of forward passes to 20 to ensure more reliable estimates.

For language models, we fine-tune pre-trained T5 models downloaded from Hugging Face. We use the Adam optimizer with momentum 0.9 and weight decay of $10^{-4}$. For summarization, we use a learning rate of $10^{-5}$, batch size of 16, and maximum generation length of 512 tokens. For translation, we use a learning rate of $10^{-4}$, batch size of 32, and a maximum generation length of 128 tokens. 
Fine-tuning language models is already computationally expensive, and incorporating multiple forward passes for MC dropout would render training impractical. Therefore, in the case of language models, we use a single forward pass and compute the PE directly from the output distribution.

Before applying \name, we first pre-train the models using standard task-specific losses (CE for classification and MSE for regression). Automatic mixed-precision training is employed for ResNet-50/ImageNet, Wide-ResNet/CIFAR-100, and the language models to reduce memory usage and accelerate training. All models and experiments are implemented in Python 3 using the PyTorch framework, and training is conducted on a single NVIDIA RTX A4000 GPU.

\section{Efficiency of \name}
\label{app:b}
\setlength{\tabcolsep}{4pt}
\begin{table*}[t]
    \caption{\name performance varying the number of forward passes for MC dropout.}
    \label{tab:efficiency}
    \centering
    \begin{tabular}{ccccccccc}
        \toprule
        \multirow{2}{*}{\textbf{Benchmark}} & \textbf{\# MC} & \multirow{2}{*}{\textbf{Error}$\downarrow$} & \multirow{2}{*}{\textbf{ECE}$\downarrow$} & \multirow{2}{*}{\textbf{uA}$\uparrow$} &  \multirow{2}{*}{\textbf{uAUC}$\uparrow$} & \textbf{Corr.} & \textbf{Wasser.} & \textbf{Time per} \\  
        & \textbf{samples}& & & & &\textbf{w/ res.}$\uparrow$ & \textbf{dist.}$\uparrow$ &\textbf{epoch}[s]$\downarrow$\\[2mm]  \hline
        & 1 &	0.47 & 0.22 & 0.81 & 0.88 & 0.68 & 0.25  & 6649 \\
        ResNet50 & 5& 0.47 & 0.22  & 0.82 & 0.89 & 0.69 & 0.26  & 7239 \\
        ImageNet & 10 &	0.46 & 0.22 & 0.82 & 0.89 & 0.69 & 0.26   & 7250\\
        & 20 &	0.46 & 0.22 & 0.82 & 0.89 & 0.69 & 0.26   & 7301\\
        \hline
        
        & 1 & 0.27  & 0.16  & 0.85 & 0.87 &	0.71& 0.19& 747 \\
        Wide-ResNet & 5   &	0.26  & 0.15 & 0.85 & 0.87 &0.72& 0.19 & 750\\
        Cifar100 & 10 &	0.26 & 0.15 & 0.85 &0.87 &	0.72	& 0.19  & 760 \\
        & 20 &	0.26 & 0.15 & 0.85 &0.87 &	0.71& 0.19  & 775 \\
        \hline
        & 1 & 0.1 & 0.02 & 0.91 & 0.91 & 0.65 & 0.34 & 231\\
        ResNet18& 5  &	0.1  & 0.01  & 0.92 &	0.93 & 0.65 & 0.34 & 235 \\
         Cifar10& 10 & 0.1 & 0.02 & 0.92 & 0.93 & 0.66 & 0.34  & 243 \\
        & 20 & 0.11 & 0.02 & 0.91 & 0.93 & 0.65 & 0.33  & 258\\ 
        \hline 
        & 1 & 0.04 & 0.02 & 0.96 & 0.88& 0.65& 0.38  & 415\\
        ResNet18& 5 & 0.04 & 0.01 & 0.96 &0.89 & 0.65 &	0.39   & 419 \\
        SVHN& 10 & 0.04 & 0.01 & 0.96 &	0.89	& 0.66 &0.4 & 430\\
        & 20 & 0.04 & 0.01 & 0.96 &	0.9	& 0.66 & 0.4 & 437\\
        \bottomrule
    \end{tabular}

    \begin{tabular}{cccccccccc}
        \toprule
         \multirow{3}{*}{\textbf{Bench.}} &  \multirow{3}{*}{\shortstack{\textbf{\# MC}\\ \textbf{samples}}} &  \multirow{3}{*}{\textbf{MSE}$\downarrow$} &  \multirow{3}{*}{\textbf{ECE}$\downarrow$} & \textbf{Corr. } & \textbf{Corr. } &  \multirow{3}{*}{\textbf{NLL}$\downarrow$}  &  \multirow{3}{*}{\textbf{ENCE}$\downarrow$} &  \multirow{3}{*}{\textbf{AUSE}$\downarrow$} & \textbf{Time } \\
        & & & & \textbf{Error} & \textbf{Error} & & &  &\textbf{per}\\
        &  & & & \textbf{PE}$\uparrow$ & \textbf{Var}$\uparrow$ & & &  &\textbf{epoch}[s]$\downarrow$ \\[2mm] 
        \hline 
        & 5 & 17.9 & 0.03& 0.27 & 0.28 & 4.48 & 1.42 & 0.05 & 2956 \\
        NYC  & 10  & 16.8 & 0.02 & 0.30 & 0.33 & 2.45 & 0.91& 0.04 & 4031 \\
        taxi & 20 & 16.9 	& 0.02 & 0.30 & 0.39 & 2.33	& 0.82 & 0.03 & 6221 \\
         & 50  & 16.4 & 0.02 & 0.29 & 0.36 & 1.89 & 0.44 & 0.03 & 13675 \\
        \hline
         \multirow{4}{*}{MSD}& 5 &119.6 & 0.1 & 0.01 & 0.01 & 7.33 & 2.08 & 0.02 & 95\\
        & 10& 119.3 & 0.09 & 0.05 &0.05 & 4.39 & 1.27 & 0.01 & 143\\
        & 20&  112.9 & 0.09 & 0.15 & 0.1 & 3.77  & 0.96 & 0.01 & 253\\
        & 50& 118.8 & 0.1 & 0.13 & 0.12 & 3.70 & 0.95 & 0.01 & 598\\
        \hline
        & 5 & 14.4 & 0.02 & 0.32 & 0.45 & 3.3 & 1.16 & 0.05 & 0.35\\
        Boston & 10& 15.1 & 0.01 & 0.36 & 0.60 & 2.79 & 0.97 & 0.03 & 0.53 \\
        Housing & 20 & 15.3 & 0.02 & 0.42 & 0.57 & 2.08 & 0.68 & 0.03 & 0.9 \\
         & 50 & 15.2 & 0.02 & 0.39 & 0.51 & 2.4 & 0.78 & 0.04 & 2.06\\
        \bottomrule
    \end{tabular}
    
\end{table*}
\raggedbottom

The main source of overhead of \name is due to MC dropout to estimate uncertainty of a prediction. 
\name estimates both predictions and uncertainties during training and inference by employing stochastic inference techniques such as MC Dropout, which aggregates outputs from multiple forward passes through the model, i.e., 
\begin{equation}
    \mathbb{P}(y|x) \approx \mathbb{P}_{MC}(y|x) = \frac{1}{K} \sum_{k=1}^{K} \mathbb{P}_{k}(y|x).
\end{equation}
where $\mathbb{P}_{k}(y|x)$ denotes the predictive distribution from the $k$-th stochastic forward pass, and $K$ is the number of MC samples.
This enables both the estimation of predictive uncertainty and an improvement in prediction quality. 
However, this introduces additional computational overhead.
To assess the trade-offs, we next evaluate the impact of varying the number of forward passes $K$ in MC Dropout on prediction error, uncertainty quality, and training time, across both classification and regression benchmarks. Table~\ref{tab:efficiency} summarizes the results.

For classification tasks, we observe that increasing the number of MC dropout samples (from 1 to 20) results in only marginal or negligible improvements across all metrics. For instance, on ImageNet, the classification error remains nearly constant, with ECE and uA metrics showing no significant variation across different sampling counts.
This stability is primarily attributed to the nature of classification outputs. NNs for classification inherently produce a probability distribution over classes (via softmax) for each forward pass. Even a single stochastic forward pass provides a approximation of the output distribution, reducing the need for additional sampling to estimate uncertainty accurately.

In contrast, regression tasks exhibit a clear dependence on the number of MC samples. Increasing the number of forward passes leads to notable improvements in both prediction accuracy and uncertainty estimation. For example, in the NYC Taxi dataset, MSE decreases from 17.9 to 16.4, while the NLL drops substantially from 4.48 to 1.89 as the number of samples increases from 5 to 50. Similar trends are observed across other regression benchmarks, with improvements in ENCE and AUSE metrics indicating better-calibrated and more informative uncertainty estimates.
This difference arises because regression models typically output a single scalar prediction. Multiple MC samples are thus essential to approximate the posterior predictive distribution and capture the associated uncertainty, which cannot be inferred from a single prediction.

Overall, our results indicate that increasing the number of MC dropout samples has limited utility in classification, where the model already produces a rich output distribution, but substantially benefits regression tasks, where uncertainty must be explicitly estimated via multiple stochastic forward passes.

\end{document}